\documentclass[twoside]{article}

\usepackage{aistats2020}
\usepackage{mathtools}
\usepackage{amssymb}
\usepackage{booktabs}
\begin{document}

%

%

\twocolumn[

\aistatstitle{Rediscovering Deep Neural Networks Through Finite State Distributions}

\aistatsauthor{Amir Emad Marvasti\And  Ehsan Emad Marvasti\And George Atia \And Hassan Foroosh}

\aistatsaddress{ aemad@cs.ucf.edu \And eemad@cs.ucf.edu \And george.atia@ucf.edu \And foroosh@ucf.edu  }
\aistatsaddress{University of Central Florida}
]
\begin{abstract}
We propose Finite Konvolutional Neural Networks (FKNN) as a glass-box, discrete counterpart of Convolutional Neural Networks, in which the linear and non-linear components of the network are naturally derived and justified in terms of Bayes' Theorem.
The building blocks of our network are classifiers operating on distributions over finite spaces. 
This property enables composition of bayesian classifiers to construct more expressive models.
The resulting composite model consists of linearities and non-linearities remarkably similar to CNNs. 
In addition to their similar functionality to CNNs, the roles of parameters, variables and layers are clear from a statistical perspective. 
Parameters and variables represent Probability Mass Functions (pmf) in FKNNs, providing the potential for usage of the arsenal of statistical and information theoretical methods.
Unlike existing designs that rely on heuristics, the proposed framework restricts subjective interpretations of Neural Networks and creates an alternative method to approach large scale machine learning. 
\end{abstract}

\section{Introduction}
The ever-increasing complexity of Convolutional Neural Networks (CNN) and their associated set of layers demand deeper insight into the internal mechanics of CNNs.
Despite the fact that the prediction layer of CNNs (e.g., the Softmax layer) and the loss functions (e.g., Cross Entropy) are borrowed from the Bayesian framework with a clear interpretation, an exact connection between the functionality of the intermediate layers with statistics remains elusive.
The current understanding of CNNs leaves much to subjective intuitions, designs and extensive experimental justifications.

We argue that subjectivity is inherent in statistical problems defined over real numbers.
Accordingly, the confusion existing in the functionality of CNNs partially corresponds to the aforementioned theoretical subjectivity.
Since real vector spaces are unbounded and uncountable, one requires strong assumptions in the form of prior information about the underlying data distribution in a Bayesian framework.
For example, fitting a Gaussian distribution to a set of samples requires that the prior distribution on the location parameter be non-vanishing near the samples.
In this scenario, an uninformative prior needs to be close to the uniform distribution over real numbers; a paradoxical distribution.
Although the choice of the prior in univariate distributions is not a practical issue, the adverse effects of subjective priors are more evident in high dimensions.
When the sample space is large and the data is comparatively sparse, either careful design of a prior or a high variance prior is needed.
Note that in the context of CNNs, the architecture, initialization, regularization, and other processes can be interpreted as imposing some form of prior on the distribution of real data.
The theoretical complications of inference over real valued data motivated this paper to investigate inference over large finite domains.

In contrast to distributions defined over real numbers, working with Finite State Distributions(FSD) makes the problem of objective inference theoretically more tractable.
Meanwhile, embedding the real valued data in large finite spaces with factored representation is expressive for practical purposes.
In problems where the data are represented by real numbers, the values can be treated as parameters of a finite-state distribution, therefore each sample represents a distribution over some finite space.
Alternatively, real valued data can be quantized with a trainable statistical model, e.g. Gaussian Mixture Models.

In this paper, we present a new deep learning framework with the key feature that unlike existing CNN models the domain of the classifier is the probability simplex.
Classification of FSDs are attractive in the sense that it sets up the requirement for composition of classifiers, since the output of Bayesian classifiers are FSDs.
The composition of Bayesian FSD classifiers are used to serve as a multilayer classification model. 
To construct a Bayesian FSD classifier we introduce the KullBack-Leibler divergence (KLD) as the log-likelihood function.
The resulting composite model deeply resembles CNNs, where modules similar to the core CNN layers are naturally derived and fit together.
\begin{itemize}

\item
We show that general forms of popular non-linearities used in deep neural networks, e.g., ReLU and Sigmoid \cite{nair2010rectified}, are derived naturally using Bayesian classification.
\item
We show that the linearities amount to calculating the KLD, while marginalization of the spatial indices is asymptotically equivalent to max pooling.
In our framework, a natural correspondence exists between types of the nonlinearity and pooling. In particular,
Sigmoid and ReLU correspond to Average Pooling and Max Pooling, respectively, while usage of each pair is dictated by the type of KLD employed.
\item
The models in our framework are statistically analyzable in all the layers; there is a statistical interpretation for every parameter, variable and layer.
All the tensors existing in our framework represent a FSD.
The interpretability of the parameters and variables provides insights into the initialization, encoding of parameters and the optimization process.
Since the distributions are over finite states, the entropy is easily calculable for both the model and data, providing a crucial tool for both theoretical and empirical analysis.
\end{itemize}
Overall, we set up the stepping stones toward understanding and interpretable design of large scale machine learning toolboxes.

The organization of the paper is as follows. In Section \ref{sec:related}, we review related work on FSDs and the analysis of CNNs.
In Section \ref{sec:frame}, we describe the construction of the proposed framework, multilayer and convolutional models for classification.
In Section \ref{sec:experiment}, we evaluate few baseline architectures in the proposed framework as a proof of concept, and provide an analysis on entropy measurements available in our framework.
\section{Related Work}\label{sec:related}
A line of work on statistical inference in finite-state domains focuses on the problem of Binary Independent Component Analysis (BICA) and the extension over finite fields, influenced by \cite{barlow1989finding,barlow1989unsupervised}.
The general methodology in the context of BICA is to find an invertible transformation of input random variables which minimizes the sum of marginal entropies \cite{yeredor2011independent,yeredor2007ica,e2011immune,painsky2014generalized,painsky2016large}.
Although the input space is finite, the search space for the correct transformation is computationally intractable for high-dimensional distributions given the combinatorial nature of the problem.
Additionally, the number of equivalent solutions is large and the probability of generalization is low.

In the context of CNNs, a body of research concerns discretization of variables and parameters of neural networks \cite{courbariauxbinarynet,soudry2014expectation,courbariaux2015binaryconnect}.
\cite{rastegari2016xnor} introduced XNOR-Networks, in which the weights and the input variables take binary values.
While discretization of values is motivated by efficiency, the optimization and learning the representation of the data are in the context of real numbers and follow similar dynamics as in CNNs.

To formalize the functionality of CNNs, a wavelet theory perspective of CNNs was considered by \cite{mallat2016understanding} and a mathematical baseline for the analysis of CNNs was established. 
\cite{tishby2000information} introduced the Information Bottleneck method (IBP) to remove irrelevant information and maintain the mutual information between two variables.
\cite{tishby2015deep} proposed to use IBP, where the objective is to minimize the mutual information between consequent layers, while maximizing the mutual information of prediction variables and hidden representations.
\cite{su2017probabilistic} introduce a framework for stochastic non-linearities where various non-linearities including ReLU and Sigmoid are produced by truncated Normal distributions.
In the context of probabilistic networks, Sum Product Networks (SPNs)~\cite{poon2011sum,gens2012discriminative,gens2013learning} are of particular interest, where under some conditions, they represent the joint distribution of input random variables quite efficiently.
A particularly important property of SPNs is their ability to calculate marginal probabilities and normalizing constants in linear time.
The efficiency in the representation, however, is achieved at the cost of restrictions on the distributions that could be estimated using SPNs.
\cite{patel2016probabilistic} constructed Deep Rendering Mixture Models (DRMM) generating images given some nuisance variables. They showed that given that the image is generated by DRMM, the MAP inference of the class variable coincides with the operations in CNNs.
\section{Proposed Framework}\label{sec:frame}
\subsection{Notation}
For every random variable $V$, we denote the range of $V$ as $R(V)$.
Throughout the paper element $i$ of a vector $v$ is represented as $v_i$. For vector $v$ and matrix $w$, $\log(v),e^v,\log(w),e^w$ represent element-wise operations of logarithm and exponentiation. 
The $d$ dimensional simplex is denoted as $\Delta^d$ while $\Delta^d_\ell =\lbrace x\in \mathbb{R}^d| \sum_{i=1}^d e^{x_d}=1\rbrace$ and its elements are referred to as the logarithmic simplex of dimension $d$ and log-pmfs of dimension $d$ respectively.
Kullback Leibler Divergence (KLD) of $p,q \in \Delta_\ell^d$ denoted as $D(p||q)$ is equal to $ \sum_{i=1}^d e^{p_i}\left(p_i- q_i\right)$.
\subsection{Input Modeling}\label{subsec:inputmodel}
We set up our framework by modeling the input data as a set of log-pmfs $\lbrace x^{(i)}\rbrace_{i=1}^n$ where $x^{(i)}\in \Delta_\ell^d $.
To demonstrate an example of interpreting real-valued data as a pmf, consider a set of $m$-pixel RGB image data.
We can view each pixel as a log-pmf over the set $\lbrace \textrm{R},\textrm{G},\textrm{B}\rbrace$ and further interpret the value of each channel as the unnormalized log-probability of being in the corresponding state.
If we normalize the log-pmf of each pixel, we can interpret the image as a factorized log-pmf over $3^m$ states, with $m$ factors and each pixel representing a log-pmf over $d=3$ states.
Formally, we define a transfer function $\Phi:\mathbb{R}^\nu \to \Delta_\ell^d$.
In the previous example, each pixel is mapped from $\mathbb{R}^3$ (i.e., $\nu=3$) to $\Delta_\ell ^3$ (i.e., $d=3$). Therefore, the entire image is mapped from $\mathbb{R}^{3m}$ to $\Delta_\ell^{(3^m)}$.
In general, the choice of $\Phi$ depends on the nature of the data and it can either be designed or considered as a trainable layer. 

\subsection{Classification}
We continue by defining two layers to perform classification of pmfs using Bayes' Theorem.
Consider the class variable $H$ with $|H|$ states and $X$ where $R(X)=\Delta_\ell^d$ we can write
\begin{align}
    P_{H|X}(h|X) = \frac{P_{X|H}(x|h)P_H(h)}{P_X(x)}\label{eq:bayes}.
\end{align}
The choice of the likelihood function $P_{X|H}(x|h)$ determines the classification procedure. 
There are two popular likelihood functions defined over the simplex, namely Multinomial distribution and Dirichlet distribution.
As shown in \cite{cover2012elements,csiszar1998method} logarithm of Multinomial distribution asymptotically converges to the KLD of empirical distribution of observations and the parameters.
Similar observation can be made in the case of Dirichlet distribution, where the KLD is reversed(Appendix A.).

Accordingly, we use the following approximations for likelihood of distribution $x\in \Delta_\ell^d$ given the distribution $q^h\in \Delta_\ell^d$ and $\alpha \geq 0$ as the parameters.
\begin{align}
P^s_{X|H}(x|h) &\approx \exp(-\alpha D(x||q^h))\label{eq:skld}\\
P^r_{X|H}(x|h) &\approx \exp(-\alpha D(q^h||x))\label{eq:rkld},
\end{align}
where $P^s$ and $P^r$ are approximations of asymptotic likelihood corresponding to Multinomial and Dirichlet distributions respectively.

We will show that approximations to ReLU-type networks and Sigmoid-type networks are derived when employing $P^r$ and $P^s$ likelihoods, respectively.
We define a single layer model for classification as an example of using $P^r$.
Constructing models with $P^s$ follows a similar derivation.

To calculate the log-probability of an input $x$ belonging to class $h$ following the Bayesian framework, we substitute (\ref{eq:rkld}) in logarithm of (\ref{eq:bayes}).
\begin{align}
\log ( P_{H|X}&(h|x)) =  -\alpha D(q^h||x) + \log(P_H(h))\nonumber\\
&- \log\left(\sum_{h'} \exp(-\alpha D(q^{h'}||x)+\log(P_H(h')))\right)
\label{eq:singlayerclass}
\end{align}

We can break the operation in (\ref{eq:singlayerclass}) into composition of a linear mapping $\textrm{Divg}^r(.)$, calculating the log-likelihoods and a non-linear mapping $\textrm{LNorm}(.)$, log normalizing the output distribution.
\begin{align}
&\textrm{Divg}^r : x\in \Delta_\ell^d,w\in \Delta_\ell^{v\times d} ,b \in \Delta_\ell^{v},\alpha \in \mathbb{R}^+ \to \mathbb{R}^v\label{eq:domainlinear},\nonumber\\
  &\Delta_\ell^{v\times d} = \lbrace w' \in \mathbb{R}^{v\times d}|(w'_{i,:})^T \in \Delta_\ell^d\rbrace
\end{align}
where $w'_{i,:}$ is the $i$-th row of the matrix $w'$. We define the function $\textrm{Divg}$ as
\begin{align}
\textrm{Divg}^r&(x;w,b,\alpha) = \alpha\left(e^wx+ \mathcal{H}(w)\right) + b \label{eq:linfuncdef}.
\end{align}
Each row of $w$ contains a distribution and $\mathcal{H}(w):\Delta_\ell^{v\times d}\to ({\mathbb{R}^+})^v$ calculates the entropy of each row. The weights $w$ and biases $b$ being the parameters of the model, are randomly initialized and trained according to some loss function. Note that the $\mathrm{Divg}^r$ is \textbf{linear} with respect to $x$.
 Unlike current CNNs, the familiar terms in (\ref{eq:linfuncdef}) such as the linear transformation $w$ and the bias term $b$ are not arbitrary.
 Specifically, $wx$ is the cross entropy of the sample and the distributions, while $b$ is the logarithm of the bias term in (\ref{eq:bayes}).
 The entropy $\mathcal{H}(w)$ can be thought as the regularizer matching the Maximum Entropy Principle \cite{jaynes1957information}.
 The $\mathcal{H}(w)$ term biases the probabilities to classes with highest degree of uncertainty in their likelihood function.

The non-linear function $\textrm{LNorm}:\mathbb{R}^v\to \Delta_{\ell}^v$ is the Log Normalization function whose $i$-th component is defined as 
\begin{align}
&\textrm{LNorm}_i(x) = x_i - \log\left(\sum_{j=1}^v e^{x_j}\right)\:.
\label{eq:logsoftdef}
\end{align}
Note that $\textrm{LNorm}(.)$ is a multivariate operation. The behavior of $\textrm{LNorm}$ in one dimension of the output and input is similar to that of \textbf{ReLU}.

When $P^s$ is used as the likelihood, the classification can be written as the composition of $\textrm{Divg}^s$ and $\textrm{ENorm}$(Exponential Normalization) where 
\begin{align}
&\textrm{Divg}^s : x\in \Delta^d,w\in \Delta_\ell^{v\times d} ,b \in \Delta_\ell^{v},\alpha \in \mathbb{R}^+ \to \mathbb{R}^v\label{eq:domainps}\\
&\textrm{Divg}^s(x;w,b,\alpha) = \alpha\left(wx+ \mathcal{H}(\log(x))\right) + b \label{eq:divgs}\:.
\end{align}
To match the domain of $\textrm{Divg}^s$ and range of normalizing function, we define $\textrm{ENorm}:\mathbb{R}^v \to \Delta^v$ as
\begin{align}
\textrm{ENorm}_i(x) =\frac{e^{x_i}}{\sum_{j=1}^v e^{x_j}}\:.
\end{align}
$\textrm{ENorm}$ behaves similar to \textbf{Sigmoid} function.
Note that $\textrm{Divg}^s$ is not linear with respect to $x$ because of the entropy term.
Furthermore, $\alpha$ in (\ref{eq:linfuncdef}) controls the uncertainty of the output of normalization layers. For example, when $\alpha=0$, equal likelihood probability is assigned to all possible input distributions, whereas when $\alpha$ is large, a slight deviation of the input from the parameter vector results in a significant decrease in the likelihood probability. We refer to $\alpha$ as the concentration parameter, however in all the models presented we fix $\alpha=1$. 
\begin{figure*}
  \centering
  \includegraphics[scale=0.2]{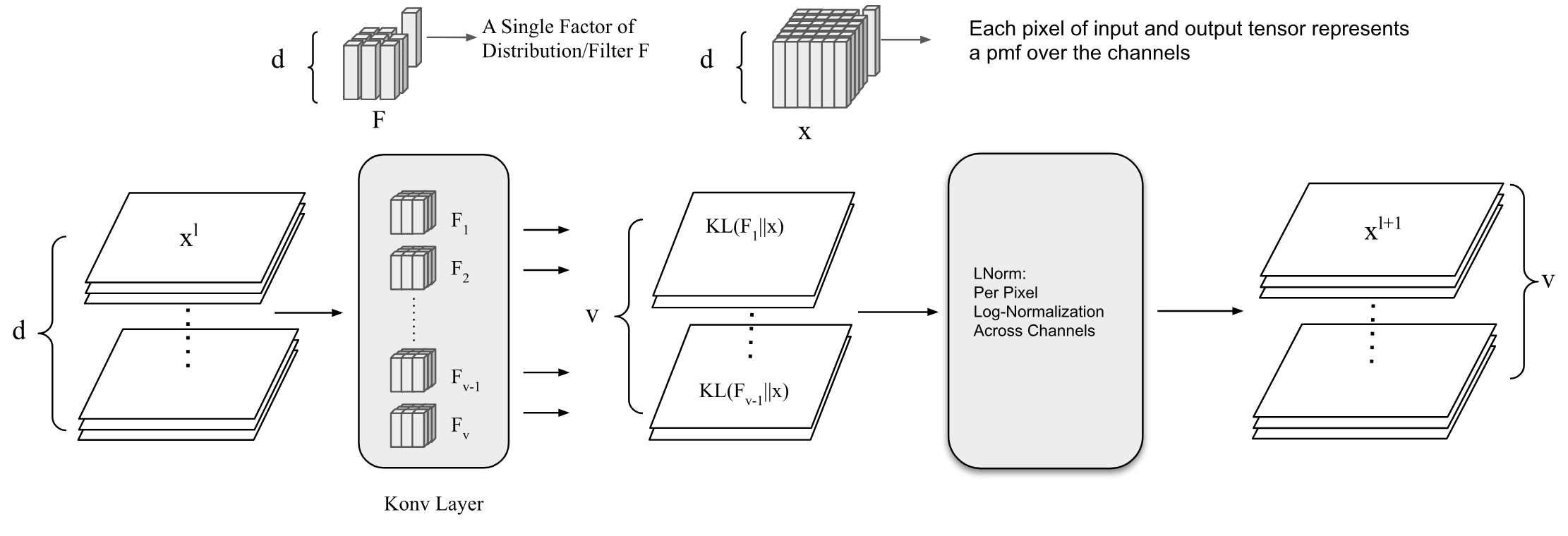}
  \caption{The Structure of the KL Convolution layer and the Normalization. The filters and input/output tensors represent factorized pmfs. The composition of these layers is equivalent to a Bayesian classifier, in which the log-likelihood is calculated by KLD and further normalized by the $\textrm{LNorm}$ layer. }\label{fig:Architecture}
\end{figure*}
\subsection{Multilayer Model}\label{sec:MultilayerConvPool}
The model described in the previous section demonstrates a potential for a further recursive generalization, i.e., the input and output of the model are both distributions on finite states. 
We extend the model simply by stacking single layer models. The input of each layer is either in $\Delta$ or $\Delta_\ell$ therefore the choice of normalization function is crucial to maintain the recursion.
The multilayer model $\textrm{FNN}(x)$ (Finite Neural Network) is defined as the composition of $\textrm{Divg}$ and $\textrm{Norm}$ layers,
\begin{align}
\textrm{FNN}(x) = \textrm{Norm}^L \circ \textrm{Divg}^{L}\circ \ldots \textrm{Norm}^l\circ \textrm{Divg}^{l} \ldots\nonumber\\ \circ \textrm{Norm}^1\circ \textrm{Divg}^1 (x) 
\end{align}
where the superscript $l$  denotes the layer index and $L$ is the total number of layers.
To elaborate, after each couple of layers, the input to the next layer are the probabilities/log-probabilities of class variable. Therefore, one can interpret the intermediate variables as distributions on a finite set of states.
\smallbreak
\subsection{Convolutional Model and Konv layer}
One of the key properties of the distribution of image data is strict sense stationarity, meaning that the joint distribution of pixels does not change with translation.
Therefore, it is desirable that the model be shift-invariant. Inspired by CNNs, we impose shift invariance functionality by convolutional KLD (\textbf{Konv}) layers.

Konv layers has a close resemblance to the Conv layers in CNNs, both in terms of parameterization and inputs.
They are parameterized with a set of filters and a bias vector, while the input is a 3 dimensional tensor.
The input tensor of the Konv layer, similar to Section \ref{subsec:inputmodel} must represent a factorized pmf and as a consequence each spatial neighborhood represents a factorized pmf. 
The filters on the other hand represent factorized pmfs where each pixel represent a single factor.
The channel size of the filers represent the number of states of each factor and should match channel size of the input tensor.
It is important to note that KLD of two factorized distributions is the sum of KLD of their individual factors.
To calculate log likelihoods locally, the filters are shifted spatially on the image tensor and KLD of neighborhoods and the filters are calculated.
This operation yields a single slice of log-likelihoods per filter which are finally stacked to form a tensor.
Finally, the tensor is normalized per pixel and across channels.

To be precise we define $\textrm{Konv}^s$ and $\textrm{Konv}^r$, corresponding to (\ref{eq:skld}) and (\ref{eq:rkld}) respectively.
\begin{align}
&\textrm{Konv}^r(x_\ell;F,B) &= \alpha(x_\ell\circledast e^F + \mathcal{H}(F)) + B\\
&\textrm{Konv}^s(x;F,B) &= \alpha(x\circledast F + \mathcal{H}^*(x)) + B
\end{align}
where $x_\ell\in \Delta_\ell$, $x\in \Delta$ and $F$ represents the set of filters in the layer (each filter is normalized and in the logarithm domain), $\mathcal{H}(F)$ is the vector of filter entropies, $\circledast$ is the convolution operator used in conventional CNNs and $\alpha \in \mathbb{R}^+$ is the concentration parameter.
$\mathcal{H}^*(x)$ is the tensor corresponding to the entropy of each neighborhood of input and can be calculated in the following manner
\begin{align}
    \mathcal{H}^*(x) = (-x\log(x))\circledast \textbf{1}(F),
\end{align}
where $\textbf{1}(F)$ is the tensor $F$ filled with ones.

Finally (E/L)Norm is applied to the output of the Konv, per pixel and across the channels.
As an example the overall operation of $\textrm{Konv}^r$ and LNorm layers is illustrated in Fig.\ref{fig:Architecture}.

\smallbreak
\noindent\textbf{Pooling:} We define the pooling function as a marginalization of indices in a random vector. In the case of tensors extracted in FNNs, the indices correspond to the relative spatial positions. In other words, the distributions in the spatial positions are mixed together through the pooling function.
Assume $x^l$ is the input to the pooling layer, where $x^l_{r,s,:} \in \Delta_\ell^V$. To calculate the marginalized distribution the input needs to be transferred to the probability domain. After marginalization over the relative spatial index, the output is transferred back to the logarithm domain.
We define the logarithmic pooling function $x^{l+1} = \textrm{LPool}(x^{l};p_{r,s})$ as
\begin{align}
x^{l+1}_{\hat{r},\hat{s},k} = \log\left(\sum_{i,j\in \textrm{supp}(p_{r,s})}\exp (x^{l}_{\hat{r}+i,\hat{s}+j,k})\, p_{r,s}(i,j)\right)\label{eq:LSEpool}, 
\end{align}
where $p_{r,s}$ is the probability distribution over the relative spatial positions and $\textrm{supp}(.)$ denotes the support. In the usual setting of pooling functions and our model, $p$ is assumed to be a uniform distribution and the support of the distribution represents the pooling window.
Note that the $\log(\sum \exp(.))$ term in (\ref{eq:LSEpool}) is \textbf{approximately equivalent to the Max function} as the variables in the exponent deviate.
Therefore, we hypothesize that Max Pooling in conventional CNNs is approximating (\ref{eq:LSEpool}). Evidently, the output of the pooling function is already normalized and is passed to the next layer.
In the case that (\ref{eq:skld}) is used, the input and output of pooling layer need to be in the probability domain, therefore the pooling function will become identical to average pooling.
\smallbreak
\noindent\textbf{Input Layer:} The model presented so far considers finite state probability distributions as input to the layers. In the case of natural images, we chose to normalize all the pixel values to the interval $(0,1)$. Each pixel value was interpreted as the expectation of a binary random variable with range $\lbrace 0,1 \rbrace$. As a result, each filter with $m$ total number of variables is a probability distribution over a space of $2^m$ states. Note that our model is not restricted by the choice of the type of input distribution. Depending on the nature of the input, the user can modify the distribution represented by filters, e.g., Normal distributions on real spaces.

\subsection{Parameterization}
The parameters of the model represent parameters of distributions which are constrained to some simplex. To eliminate the constraints of the parameters, we use a \textit{Link Function}, $\psi : \mathbb{R}^D \to \Delta^D$, mapping the \textit{Seed} parameters to the acceptable domain of parameters, i.e., logarithmic/probability simplex.
The link function impacts the optimization process and partially reflects the prior distribution over the parameters. While the parameters are updated in $\mathbb{R}^D$ uniformly, the mapped parameters change according to the link function.
The filters in our model are factorized distributions and each component is a categorical distribution. Additionally, the biases are categorical distributions, therefore we use similar parameterization for biases and filter components. 
In general, the filters of the model are obtained by
\begin{align}
F_{r,s,:} &= \psi(\theta_{r,s,:})\\
B &= \psi(\beta)
\end{align}
where $\theta_{r,s,:}$ are the seed parameters of the filter in the spatial position $r,s$ and across all the channels, $F_{r,s,:}\in \Delta_\ell^D$ represents the channels of the filter in the $r,s$ position, $\beta \in \mathbb{R}^V$ is the seed parameter of bias and $B\in \Delta_\ell^V$ is the bias vector.
Since the filters and biases comprise categorical distributions, we avoid complicating the notation by limiting the discussion to the parameterization of categorical distributions.
We suggest two forms of parameterization of a categorical distribution $q \in \Delta^d$, namely log-simplex and spherical parameterizations.

\subsubsection{Log-Simplex Parametrization} 
We define the link function with respect to the natural parametrization of a categorical distribution, where the seed parameters are interpreted as unnormalized log probabilities. Therefore, the ENorm as the link function
\begin{align}
q_i = \frac{e^{\theta_i}}{\sum_{j=1}^d e^{\theta_j}},\label{eq:linklogsimplex}
\end{align}
where $\theta$ is the seed parameter vector and the subscript denotes the index of the vector components.
Having the Jacobian of (\ref{eq:linklogsimplex})
\begin{align}
\frac{\partial q_i}{\partial\theta_i} = q_i(1-q_i),\quad \frac{\partial q_i}{\partial\theta_j} = -q_jq_i, \quad i\neq j,
\end{align}
we can observe that the Jacobian only depends on $q$  and does not depend on the denominator in (\ref{eq:linklogsimplex}). Therefore, the link function is invariant to translation of $\theta$ along the vector $(1,1,\ldots,1)$.
Log-Simplex parameterization completely removes the effect of the additional degree of freedom.
\smallbreak
\noindent\textbf{Initialization:}
We initialize each factor of the filters by sampling from a Dirichlet distribution with parameters equal to $1$. Therefore, the distribution's components are generated uniformly on some simplex.
We speculate that the initialization of the model should follow maximizing the mixing entropy or Shannon-Jensen Divergence (JSD) of the filters in a given layer $\lbrace q ^h\rbrace_{h=1}^{|H|}$, defined as $\Delta \mathcal{H}$
\begin{align}
\Delta \mathcal{H} = \mathcal{H}\left(\sum_{i=1}^{|H|}P_H(h) q^h\right) - \sum_{h=1}^{|H|} P_H(h) \mathcal{H}(q^h)\:,\label{eq:mixingentropydef}
\end{align}
where $|H|$ is the total number of filters, $q^h$ is the $h$-th filter and $P_H(h)$ is the corresponding bias probability.
There is a parallel between orthogonal initialization of filters in conventional CNNs and maximizing $\Delta  \mathcal{H}$ in Konv networks. In the extreme case where filters are degenerate distributions on unique states and the filters cover all possible states, $\Delta \mathcal{H}$ is at the global maximum and the Konv operation is invertible. Similarly, orthogonal initialization of conventional CNNs is motivated by having invertible transformations to help with the information flow through the layers.
Since it is hard to obtain a global maximizer for JSD, we minimize the entropy of individual filters (second term in (\ref{eq:mixingentropydef}) by scaling the log-probabilities with a factor $\gamma > 1$. We set $\gamma=\log(\#\textrm{filters})$ as a rule of thumb.
Finally, the Bias seed components are initialized with zeros, indicating equal proportion of mixture components.
\begin{table*}[t]
\caption{Accuracy of CNN architectures and their finite state counterparts on CIFAR-10 and CIFAR-100 over 150 epochs of training. The performance of the FKNNs with Log Simplex and Spherical parameterization are compared. NIN and VGG FKNNs outperform their corresponding vanilla CNN architectures. FKNNs were trained without regularization, Dropout, BatchNorm and data preprocessing. In all FKNNs, the learning rate is fixed to 1. Dropout, BatchNorm and Data Preprocessing are abbreviated DO, BN and DPP, respectively. In the case of Quick-CIFAR and VGG, FKNNs outperform their CNN counterparts when stripped of BN and DO.}
\begin{center}
\begin{tabular}{lccccccccc}
\toprule
\multicolumn{1}{c}{} &\multicolumn{2}{c}{CIFAR100}&\multicolumn{2}{c}{CIFAR10}&\multicolumn{1}{c}{}\\
\cmidrule(r{5pt}){2-3}\cmidrule{4-5}
\multicolumn{1}{c}{Model} &\multicolumn{1}{c}{test(\%)} &train(\%)&test(\%)&train(\%)&params&DO&BN&DPP \\
\cmidrule(r{5pt}){1-1}\cmidrule(r{5pt}){2-3}\cmidrule(r{5pt}){4-4}\cmidrule(r{5pt}){5-5}\cmidrule(r{5pt}){6-6}\cmidrule(r{5pt}){7-9}
\multicolumn{1}{l}{Quick-CIFAR-FKNN-Spherical}  & \textbf{41.14}&43.62&\textbf{78.42}&78.98&0.1M& & &  \\
\multicolumn{1}{l}{Quick-CIFAR}  & 40.21&45.22&77.21&83.46&0.1M& & &\checkmark \\
\multicolumn{1}{l}{Quick-CIFAR-FKNN-LogSimplex}  & 38.28&39.28&72.76&72.97&0.1M& & &  \\

\midrule
\multicolumn{1}{l}{NIN} &57.47&81.59&84.69&96.06&1.5M&\checkmark&&\checkmark\\
\multicolumn{1}{l}{NIN} &\textbf{49.50}&99.04&\textbf{81.28}&99.36&1.5M&&&\checkmark\\
\multicolumn{1}{l}{NIN-FKNN-Spherical}  &48.92&54.13&80.63&86.03&1.5M&&& \\
\multicolumn{1}{l}{NIN-FKNN-LogSimplex}  &42.33&44.95&75.30&77.28&1.5M& & &  \\

\midrule
\multicolumn{1}{l}{VGG}  & 68.75 & 98.59 & 91.41 & 99.30 &14.7M&\checkmark&\checkmark&\checkmark \\
\multicolumn{1}{l}{VGG-FKNN-LogSimplex} & \textbf{51.77}&63.93&\textbf{83.16}&91.49&14.7M& & &\\
\multicolumn{1}{l}{VGG-FKNN-Spherical} & 33.15&36.04&68.86&69.07&14.7M& & &\\
\multicolumn{1}{l}{VGG}  &37.03&99.96   & 76.15 & 100 &14.7M&&&\checkmark \\

\bottomrule
\end{tabular}
\end{center}
\label{Tab:ResQuick}
\end{table*}
\subsubsection{Spherical Parameterization} Here, we present an alternative parameterization method attempting to eliminate the learning rate hyper-parameter.

Assume that we parameterize the categorical distribution $q$ by the link function 
\begin{align}
q_i = \frac{\theta_i^2}{\sum_{j=1}^d \theta_j^2}\label{eq:linkpolar}\:.
\end{align}

The expression in (\ref{eq:linkpolar}) maps $\theta$ to the unit sphere ${\cal S}^{d-1}\subset \mathbb{R}^d$, where the square of the components are the probabilities. The mapping defined in (\ref{eq:linkpolar}) ensures that the value of the loss function and predictions are invariant to scaling $\theta$. The Jacobian of (\ref{eq:linkpolar}) is
\begin{align}
\frac{\partial q_i}{\partial \theta_i} &= 2\sqrt{\frac{q_i}{\lVert\theta\rVert^2}}(1- q_i)\textrm{sign}(\theta_i),\nonumber\\
\frac{\partial q_i}{\partial \theta_j} &= 2\sqrt{\frac{q_j}{\lVert \theta \rVert^2}}(- q_i)\textrm{sign}(\theta_i), \quad i\neq j.\label{eq:jacob}
\end{align}
It is evident from (\ref{eq:jacob}) that the norm of the gradient has an inverse relation with $\lVert \theta \rVert$.
Scaling $\theta$ is equivalent to changing the step size, since the direction of gradients does not depend on $\lVert \theta \rVert$.
Additionally, the objective function is not dependent on $\lVert \theta \rVert$, therefore the gradient vector obtained from the loss function is orthogonal to the vector $\theta$. Considering the orthogonality property, updating $\theta$ along the gradients always increases the norm of the parameter vector. As a consequence, the learning rate decreases at each step of the iteration; independent of the network structure.
\smallbreak
\noindent\textbf{Initialization:} The seed parameters are initialized uniformly on ${\cal S}^{d-1}$. The standard way of generating samples uniformly on ${\cal S}^{d-1}$ is to sample each component from a Normal distribution $\mathcal{N}(0,1)$ followed by normalization.

\section{Experimental Evaluations}\label{sec:experiment}
We experimented with the finite framework using the CIFAR-10 and CIFAR-100 datasets \cite{krizhevsky2009learning} on the problem of classification. We employed three types of base-CNN architectures, namely Quick-CIFAR, Network in Network (NIN) \cite{lin2013network} and VGG \cite{simonyan2014very,liu2015very} to experiment with different network sizes.
The CNN architectures were transformed to their \textbf{FKNN} versions by replacing Conv/ReLU/Pool with Konv/Norm/LPool.
We excluded certain layers from our transformation, e.g., the Dropout and the batch normalization (BatchNorm) \cite{ioffe2015batch}, since we have not yet developed a compatible counterpart of the layers with a solid justification for our framework.
The inputs for the original architectures were whitened, while the whitening procedure was not applied for testing FKNNs. 
In the FKNNs, We did not use weight decay \cite{krizhevsky2012imagenet}, regularization, and the learning rate was fixed to $1$ .
FKNNs were parameterized with log-simplex and spherical schemes for comparison.
Additionally, the original optimized learning rates were used to train the CNNs. The weights in all the models were regularized with $\ell_2$ norm penalty, where in the case of NIN and VGG the regularization coefficient is defined per layer.

Experiments with $\textrm{Konv}^s$ was excluded, since they achieved lower accuracy compared to $\textrm{Konv}^r$.
We justify this observation by considering two facts about $\textrm{Konv}^r$: 1) Since the input is in the probability domain, the nonlinearity behaves similar to Sigmoid, therefore the gradient vanishing problem exists in $\textrm{Konv}^s$. 2) As opposed to $\textrm{LNorm}$, $\exp(\textrm{LNorm}(.))$ is not convex  and interferes with the optimization process.
\begin{figure*}
  \centering
  \includegraphics[scale=0.3]{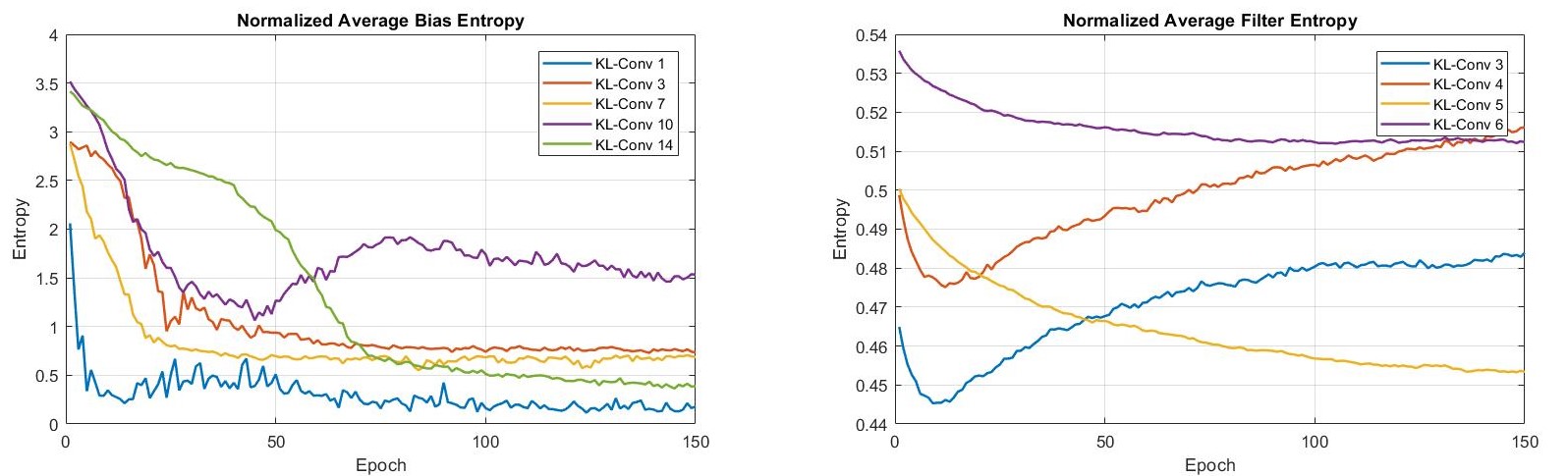}
  \caption{Entropy of bias distributions and filter distributions in the VGG-FKNN-LogSimplex network. The entropy of the bias distribution relates to the number of filters contributing to prediction. Trend of the filter entropies is roughly related to the average entropy of the initial values and the layer position in the network.}
  \label{fig:entropylayers}
\end{figure*}
Table \ref{Tab:ResQuick} demonstrates the performance achieved by the baselines and their FKNN analogues.
To provide a full comparison of our vanilla networks with their counterparts, we experimented with two setups of CNNs, i.e. with BatchNorm-Dropout and vanilla.
As seen in Table \ref{Tab:ResQuick} FKNNs outperform their CNN counterparts stripped of BatchNorm and Dropout in the case of VGG and Quick-CIFAR networks. Meanwhile the test performance of both frameworks in the NIN architecture is on a similar scale.
However, the CNN suffer heavily from overfitting where FKNNs are prone to underfitting.
This observation is explained by the representation power of both frameworks. 
FKNNs constrain both variables and parameters to the probability simplex whereas CNNs are only restricted by the form of regularization.
The representation power of FKNNs can be increased by representing each spatial position by a factorized distribution.
However, increasing the representation power of FKNNs requires further understanding of priors/initialization, which is intended for future research.


FKNNs achieved lower performance in VGG and NIN architectures which are equipped with Dropout and BatchNorm.
The results show that the finite state models' performance is at the same scale as CNNs, considering the simplicity of FKNNs.
Spherical parameterization performs better than Log Simplex in NIN-Finite and Quick-CIFAR-Finite networks, whereas in VGG-Finite Log Simplex is superior.
We do not have a definite explanation for the difference in performance of parameterizations in different architecture settings.
However, the results show that none are objectively superior as they stand.


\subsection{Entropy of Filters and Biases}
To analyze the behavior of the networks, we performed a qualitative analysis on the trend of the bias entropies and the filter entropies.
In our experiments, $\textrm{Konv}^r$ was used as the linearity. Since the input is represented by log-probability in the cross entropy term of $\textrm{Konv}^r$, the filter distribution naturally tends to low entropy distributions. However, in Figure \ref{fig:entropylayers},  we observe that the average entropy of some layers starts to increase after some iterations. This trend is visible in the early layers of the networks. Since high entropy filters are more prone to result in high divergences when the input distribution is low-entropy (property of (\ref{eq:rkld})), the network learns to approach the local optimum from low entropy distributions. The entropy of the input tensors of late layers are larger compared to that of the early layers, and start decreasing during the learning process. Therefore, the entropy of the filters decreases as the entropy of their input decreases.

The entropy of bias distributions contain information about the architecture of networks. Note that the bias component is the logarithm of the mixing coefficients. Degeneracy in the bias distribution results in removing the effect of the corresponding filters from the prediction. The increase in the entropy of the biases could also demonstrate the complexity of the input, in the sense that the input distribution cannot be expressed with a mixture of factorized distributions given the current number of filters.

\section{Conclusion}
 Our work was motivated by the theoretical complications of inference in real state spaces. 
 We argued that in finite states objective inference is theoretically feasible, while factorized finite spaces are complex enough to express the data in high dimensions.
 The stepping stones for inference in high dimensional finite spaces were provided in the context of bayesian classification.
 
 The recursive application of Bayesian classifiers resulted in FKNNs; a structure remarkably similar to Neural Networks in the sense of activations (ReLU/Sigmoid) and the linearity. 
 Consequently, by introducing the shift invariance property (Strict Sense Stationarity assumption) using the convolution tool, FKNNs as finite state analogue of CNNs were produced. The pooling function in FKNNs was derived as marginalizing the spatial position variables. The Max Pool function was explained as an approximation to the marginalization of spatial variables in the log domain. In our work, it is evident that a correspondence exists between $\textrm{Konv}^r$, ReLU and Max Pool and similarly between $\textrm{Konv}^r$, Sigmoid and Average Pool. 
 Individual layers were constructed using bayes' theorem but the overall functionality of FKNNs is not probabilistic.
 Probabilistic versions of FKNNs can be constructed by sampling from the intermediate variables, candidating FKNNs as the deteriministic backbone architecture of probabilistic models.
In the context of classic CNNs, diverse interpretations for layers and values of the feature maps exist whereas in FKNNs the roles of layers and the nature of every variable is clear. Additionally, the variables and parameters represent distributions, making the model ready for a variety of statistical tools, stochastic forward passes and stochastic optimization.
\bibliography{ref}
\end{document}